\documentclass{article}
\usepackage{spconf}

\usepackage{graphicx}
\usepackage{amssymb}
\usepackage{amstext}
\usepackage{amsmath}
\usepackage{cite}
\usepackage{hyperref} 
\usepackage{balance}

\usepackage{float}
\usepackage{ragged2e}

\title{Generating near-infrared facial expression datasets\\with dimensional affect labels}
\name{Calvin Chen, Stefan Winkler$^*$\thanks{* S.\ Winkler is now also with Asus AICS.}}
\address{School of Computing\\National University of Singapore}
\begin{document}
%
\maketitle
\begin{abstract}
Facial expression analysis has long been an active research area of computer vision. Traditional methods mainly analyse images for prototypical discrete emotions; as a result, they do not provide an accurate depiction of the complex emotional states in humans. Furthermore, illumination variance remains a challenge for face analysis in the visible light spectrum. To address these issues, we propose using a dimensional model based on valence and arousal to represent a wider range of emotions, in combination with near infra-red (NIR) imagery, which is more robust to illumination changes. Since there are no existing NIR facial expression datasets with valence-arousal labels available, we present two complementary data augmentation methods (face morphing and CycleGAN approach) to create NIR image datasets with dimensional emotion labels from existing categorical and/or visible-light datasets.  Our experiments show that these generated NIR datasets are comparable to existing datasets in terms of data quality and baseline prediction performance.
\end{abstract}
\section{Introduction}
\label{sec:intro}

Facial expression is one of the most important non-verbal channels for conveying emotions. Facial expression recognition systems based on images in the visible light (VL) spectrum have achieved a significant level of success, in large part due the availability and accessibility of large facial expression databases of VL images. However, illumination still remains a challenge for facial expression analysis. 

Adini et al.\ \cite{C3} explored the effect of illumination changes on face recognition and found that lighting conditions often change the appearance of the face in an undesirable manner. Among multiple alternatives to tackle such this limitation, the use of near infra-red (NIR) imaging seems to be a promising alternative since it can be made more robust to illumination changes.  For example, Li et al.\ \cite{C4} developed an active near infra-red (NIR) imaging system to capture face images in good condition regardless of illumination conditions. 

Besides the illumination problem mentioned above, the lack of expressivity of existing emotion datasets is also a concern. Classification of basic prototypical facial expressions has been a popular research topic. Many researchers work with the common concept of six basic emotions, which are anger, disgust, fear, happiness, sadness, and surprise \cite{Ekman}. As most facial expression databases also cover these discrete emotions, this approach has made significant strides  \cite{discrete-1,discrete-2}. However, they are necessarily limited due to the coarse granularity of emotion labels and do not work well in real-world scenarios \cite{FEA}.

\begin{figure}[htb]
\begin{minipage}[b]{1.0\linewidth}
  \centering
  \centerline{\includegraphics[width=\columnwidth]{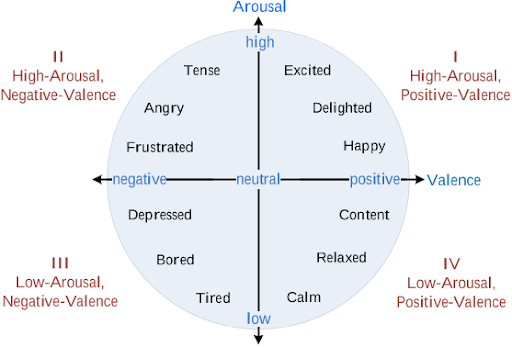}}
\end{minipage}
\caption{Circumplex model of affect \cite{circumplex}.}
\label{fig:circumplex_model}
\end{figure}

Russell \cite{russell} proposed the circumplex model as an alternative, dimensional representation of emotions. Figure \ref{fig:circumplex_model} shows the circumplex model, which maps emotions to a two-dimensional Valence-Arousal space, where valence indicates how positive or negative an emotion is, and arousal distinguishes active from passive emotions. Recently, dimensional models of affect have gained traction as they better represent subtle emotions appearing in everyday situations \cite{affectnet, aff-wild}. 

To address the issues of illumination variance and the lack of emotion granularity, we investigate using NIR datasets together with the circumplex model of emotions. However, no such datasets are available.  We therefore propose two complementary frameworks to synthesize datasets that fulfill our requirements.  The first is based on \cite{morphset} and uses facial morphing to transform categorical emotion datasets into dimensional ones. The second is based on CycleGAN \cite{cyclegan} and allows us to generate NIR images from VL datasets.  We show that results of a baseline model trained and tested on our synthetic NIR datasets are comparable to those for existing VL datasets.




\section{Datasets}
\label{sec:dataset}

\subsection{Oulu-Casia NIR\&VL dataset}

%
The Oulu-CASIA NIR\&VL facial expression database \cite{oulu-casia} consists of face images of 80 people between 23 and 58 years displaying six emotions (surprise, happiness, sadness, anger, fear and disgust). 
Both NIR and VL cameras were used to capture the same scene and expression. Images with neutral expression are not readily available and were extracted from the original video sequences.

\subsection{AffectNet}

%

AffectNet \cite{affectnet} is a database of facial expressions in the wild, created by collecting about 400,000 face images collected from the Internet.  The images were manually annotated with the basic categorical labels as well as valence/arousal. 

\subsection{VL MorphSet}

%

MorphSet  \cite{morphset} is an augmentation framework that uses facial morphing to augment categorical emotion datasets to cover a wide range of valence-arousal levels.  The resulting  dataset was created from a combination of 3 categorical datasets that have been used extensively in psychology \cite{Radboud,Karolinska,Warsaw} and consists of about 300,000 VL images of posed facial expressions with dimensional valence-arousal labels. 

\section{Methodology}
\label{sec:methodology}

We present two complementary frameworks to synthesize NIR facial expression datasets with dimensional labels.  The first uses facial morphing to augment categorical emotion datasets to cover a wide range of valence-arousal levels. The second uses CycleGAN \cite{cyclegan} to generate NIR images from VL datasets.  

\subsection{Augmentation (Categorical to Dimensional)}
\begin{figure}[H]
\begin{minipage}[b]{1.0\linewidth}
  \centering
  \centerline{\includegraphics[width=8.5cm]{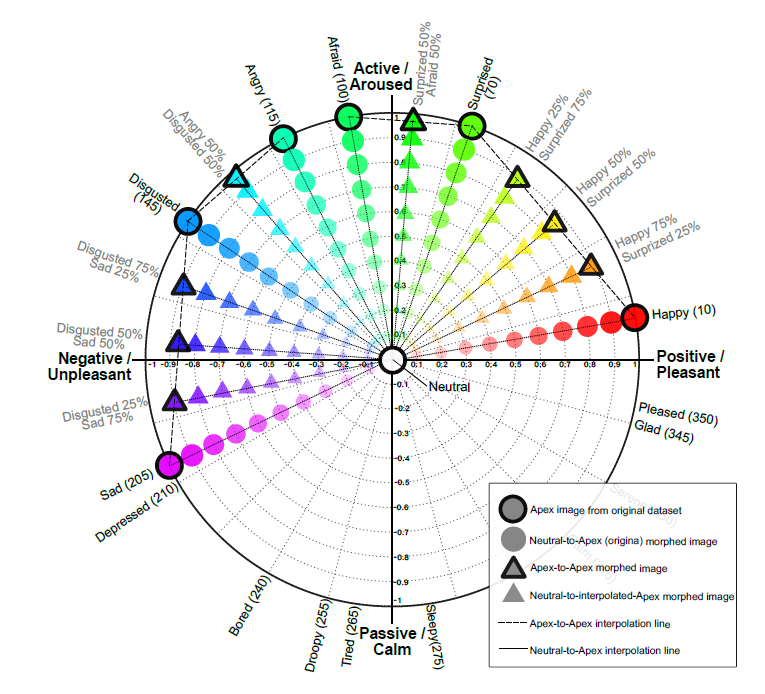}}
\end{minipage}
\caption{MorphSet augmentation framework \cite{morphset}.}
\label{fig:framework}
\end{figure}

Existing categorical datasets can be converted to a dimensional dataset with valence-arousal labels. We follow the augmentation framework proposed in \cite{morphset}. Figure~\ref{fig:framework} illustrates the augmentation process for a categorical dataset with 7 prototypical expressions. 

\begin{figure*}[t!]
    \centering
    \includegraphics[width=\textwidth]{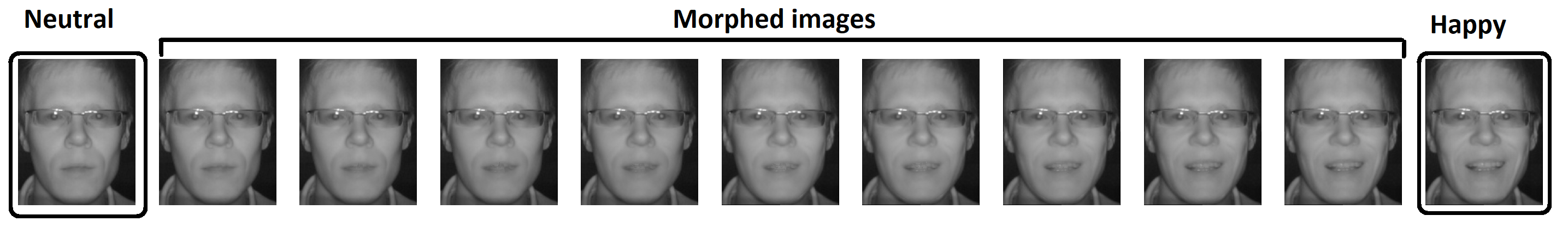}\\
    \includegraphics[width=\textwidth]{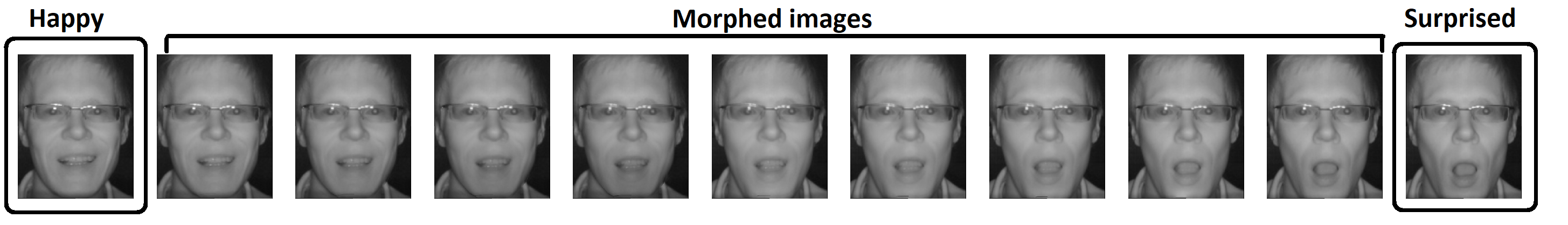}
    \caption{NIR face image morphings. Top row: Neutral-to-apex (happy) morphing. Bottom row: Apex-to-apex morphing between 'Happy' and 'Surprised' expressions. Only the left- and right-most images come from the original NIR dataset, while the remaining images are interpolated by face morphing.}
    \label{fig:morphings}
\end{figure*}

We assume a 2-dimensional Valence-Arousal (VA) space, with Neutral located at the center. The prototypical facial expressions can be mapped to points with specific coordinates in the polar AV space. The template is bounded within [10$^{\circ}$, 205$^{\circ}$] due to the lack of prototpyical emotions outside this range. We use an angle increment of 15$^{\circ}$ with an intensity increment of 0.1 as suggested in the original paper. This granularity allows us to generate up to 134 new expressions per subject. Morphed images are synthesized by Delaunay triangulation followed by local warping of 68 facial landmarks from Dlib \cite{dlib} face recognition system. Figure~\ref{fig:morphings} shows the morphed NIR face images for two example pairs of facial expressions.

Using this approach, we construct a new dimensional NIR expression dataset called \textbf{Oulu-Casia NIR MorphSet} from the Oulu-Casia NIR dataset.  With the augmentation factors described above, we could theoretically obtain $80 \times 141 = 11,280$ images. However, due to feature misalignment issues for some of the images, the final NIR MorphSet only contains $11,120$ images. Similarly, we generated an \textbf{Oulu-Casia VL MorphSet}  for reference. 

\subsection{Synthetic NIR (VL to NIR)}

We can also adapt an existing dimensional VL dataset to a dimensional NIR dataset using image-to-image translation.   CycleGAN \cite{cyclegan} is a popular architecture for such a task and works well because of its cycle-consistency loss.  We use the existing Oulu-Casia VL-NIR dataset, which contains paired images of the same subject in both VL and NIR domains, to train the CycleGAN model. Since only the infra-red intensity is captured for NIR images, they are  displayed as greyscale images. The CycleGAN model can then learn a mapping from color-to-greyscale to generate a synthetic NIR image. 

\begin{figure}[h]
\begin{minipage}[b]{1.0\linewidth}
  \centering
  \centerline{\includegraphics[width=8.5cm]{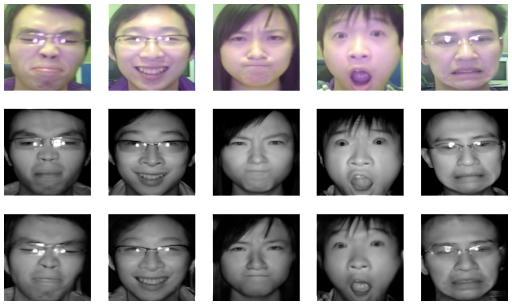}}
\end{minipage}
\caption{CycleGAN image-to-image translation. Original VL (top row) and NIR (middle row) images used for training; bottom row: synthetic NIR images produced by CycleGAN.}
\label{fig:cyclegan_train}
\end{figure}

Figure~\ref{fig:cyclegan_train} shows a sample of the images used to train our CycleGAN model and the synthetic NIR images generated by our trained model. We can see that the synthetic NIR images look visually convincing and closely resemble the original NIR images from the Oulu-Casia dataset.

\begin{figure}[H]
\begin{minipage}[b]{1.0\linewidth}
  \centering
  \centerline{\includegraphics[width=8.5cm]{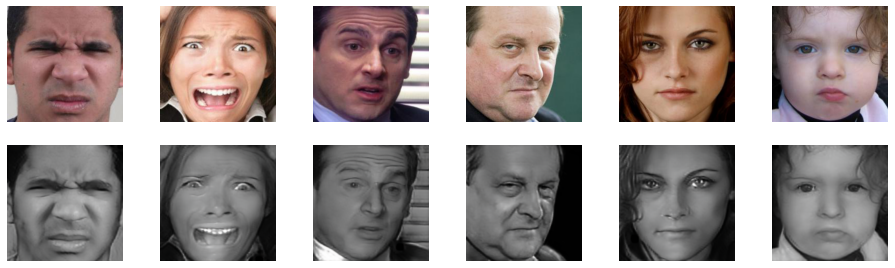}}
\end{minipage}
\caption{Original VL (top) and synthetic NIR (bottom) images generated from AffectNet.}
\label{fig:cyclegan_affectnet}
\end{figure}

\begin{figure}[H]
\begin{minipage}[b]{1.0\linewidth}
  \centering
  \centerline{\includegraphics[width=8.5cm]{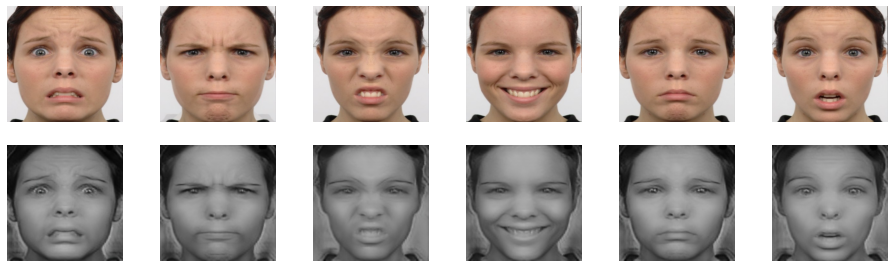}}
\end{minipage}
\caption{Original VL (top) and synthetic NIR (bottom) images generated from VL MorphSet.}
\label{fig:cyclegan_morph}
\end{figure}

We can then apply our trained VL-to-NIR CycleGAN to other VL emotion datasets.  
Figures \ref{fig:cyclegan_affectnet} and \ref{fig:cyclegan_morph} show the results for AffectNet and VL MorphSet respectively. We see that the synthetic NIR images generally look convincing, especially those from VL MorphSet. This is  because both the Oulu-Casia dataset and VL MorphSet contain frontal face images taken in a lab setting, whereas AffectNet images are completely unconstrained. 

\subsection{Baseline Model}

We use the EfficientNet-B0 \cite{efficientnet} architecture as our baseline model for facial expression recognition in valence-arousal space.  We add two neurons after the final fully connected layer in the existing EfficientNet-B0 architecture to predict valence and arousal. The $\tanh$ function is further used to constrain valence-arousal values in the range $[-1,1]$.  L2 loss is used for training.

\section{Results and Discussion}
\label{sec:results}

\subsection{Dataset Comparison}

We evaluate our baseline model’s performance on 10\% of randomly selected identities of each dataset. 

Table \ref{table:baseline_others} shows the Root Mean Square error (RMSE) and Concordance Correlation Coefficient (CCC) for each dataset. While the results are not  directly comparable across datasets due to the differences in test set, they serve as a good reference. 

\begin{table}[htb]
\centering\small
\caption{Baseline results for different datasets.}
\label{table:baseline_others}
\begin{tabular}[t]{lcccc}\hline
& RMSE & RMSE & CCC & CCC \\
& Arousal & Valence & Arousal & Valence \\ \hline
AffectNet (VL)&0.356&0.398&0.434&0.569\\
AffectNet \\ (VL-to-NIR) &0.373&0.447&0.426&0.527\\
\hline
VL MorphSet&0.124&0.107&0.880&0.945\\
VL MorphSet \\(VL-to-NIR) &0.113&0.100&0.900&0.954\\
\hline
Oulu-Casia VL \\ MorphSet&0.205&0.189&0.620&0.807 \\
Oulu-Casia NIR \\ MorphSet& 0.202&0.242& 0.592& 0.731 \\
\hline
\end{tabular}
\end{table}%

The synthetic NIR datasets -- AffectNet (VL-to-NIR) and VL MorphSet (VL-to-NIR) -- have comparable performance to their VL counterparts. This suggests that the augmentation method via CycleGAN is successful in the sense that our model is able to learn from the synthetic NIR images equally  well as from the original VL images. 

The Oulu-Casia NIR MorphSet has a similar performance to the Oulu-Casia VL MorphSet, but a worse performance as compared to the VL MorphSet despite using the same face morphing augmentation method, especially in terms of valence. This could be due to the inherent difference in datasets such as images in the VL MorphSet being less varied as compared to the Oulu-Casia NIR\&VL MorphSet.

Note that while our methods remove the need for an original NIR dimensional dataset, augmentation methods do require large amounts of data to achieve good performance, e.g.\ landmark annotations for face morphing, or paired VL\&NIR images for CycleGAN training. The data determines how well the approaches generalize to real-world scenarios.


\subsection{Error Visualization}

\begin{figure}[htb]
\begin{minipage}[b]{.48\linewidth}
  \centering
  \centerline{\includegraphics[width=4.0cm]{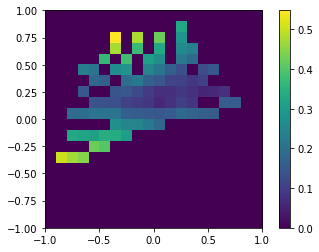}}
  \centerline{RMSE Arousal\medskip}
\end{minipage}
\hfill
\begin{minipage}[b]{0.48\linewidth}
  \centering
  \centerline{\includegraphics[width=4.0cm]{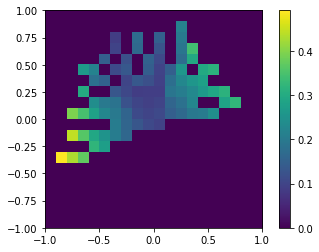}}
  \centerline{RMSE Valence\medskip}
\end{minipage}
\caption{Test RMSE of Oulu-Casia NIR MorphSet.}
\label{fig:nir_morphset_rmse}
\end{figure}

\begin{figure}[htb]
\begin{minipage}[b]{.48\linewidth}
  \centering
  \centerline{\includegraphics[width=4.0cm]{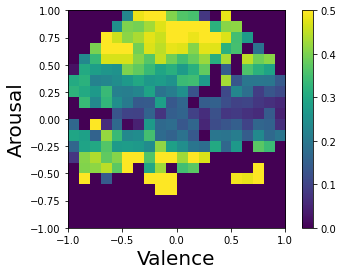}}
  \centerline{RMSE Arousal\medskip}
\end{minipage}
\hfill
\begin{minipage}[b]{0.48\linewidth}
  \centering
  \centerline{\includegraphics[width=4.0cm]{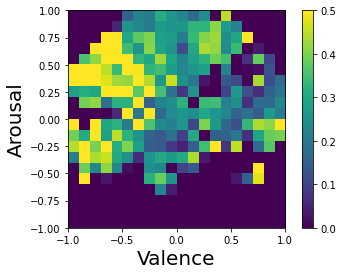}}
  \centerline{RMSE Valence\medskip}
\end{minipage}
\caption{Test RMSE of AffectNet (NIR).}
\label{fig:affectnet_rmse}
\end{figure}

Figures~\ref{fig:nir_morphset_rmse} and \ref{fig:affectnet_rmse} show the RMSE between the predicted arousal/valence and the ground-truth for the Oulu-Casia NIR MorphSet and AffectNet (NIR) respectively. Generally, the RMSE is much higher for AffectNet as compared to Oulu-Casia NIR MorphSet. However, we notice that the regions with higher error rates are similar for both datasets. This is likely due to the lack of training data for the valence-arousal values at these regions.

\balance
\section{Conclusion}
\label{sec:conclusion}

We investigated the feasibility of facial expression recognition in the near infra-red domain using the circumplex model of emotion. This is a challenging task due to the absence of datasets.

We presented two data augmentation methods to create near infra-red (NIR) datasets with dimensional valence-arousal annotations. The first method uses face morphing algorithms to transform a typical categorical dataset of facial expressions into a balanced augmented one. The second method uses CycleGAN to learn a mapping from standard VL images to NIR images. 

A deep neural network baseline was examined to classify the facial expression images and predict valence/arousal values. Our experiments show that results on our synthetic NIR datasets are comparable to results in existing datasets. We hope our findings can promote the use of NIR imagery for facial expression analysis tasks.

\newpage
\bibliographystyle{IEEEbib}
\bibliography{strings,refs}

\end{document}